\documentclass[sigconf]{acmart}
\renewcommand\footnotetextcopyrightpermission[1]{} 
\usepackage{booktabs} 
 \usepackage{mathrsfs}
\usepackage{algorithm}           
\usepackage{algorithmic} 
\usepackage{amsthm}
\usepackage{bbm}
\usepackage{color}

\newtheorem{theorem}{Theorem}

\usepackage{subfigure}

\acmConference[]{}{}{}%

\setcopyright{none}

\begin{document}
%
\title{Kernel-based Multi-Task Contextual Bandits in  Cellular Network Configuration}

\author{Xiaoxiao Wang}
\affiliation{%
  \institution{University of California, Davis}
  \streetaddress{}
  \city{Davis}
  \state{CA, USA}
  \postcode{}
}
\email{xxwa@ucdavis.edu}

\author{Xueying Guo}
\affiliation{%
  \institution{University of California, Davis}
  \streetaddress{}
  \city{Davis}
  \state{CA, USA}
  \postcode{}
}
\email{guoxueying@outlook.com}

\author{Jie Chuai}
\affiliation{%
  \institution{Noah's Ark Lab, Huawei Technologies}
  \streetaddress{}
  \city{Hong Kong}
  \state{China}
  \postcode{}
}
\email{chuaijie@huawei.com}
\author{Zhitang Chen}
\affiliation{%
  \institution{Noah's Ark Lab, Huawei Technologies}
  \streetaddress{}
  \city{Hong Kong}
  \state{China}
  \postcode{}
}
\email{chenzhitang2@huawei.com}
\author{Xin Liu}
\affiliation{%
  \institution{University of California, Davis}
  \streetaddress{}
  \city{Davis}
  \state{CA, USA}
  \postcode{}
}
\email{xinliu@ucdavis.edu}


\begin{abstract}


Cellular network configuration plays a critical role in  network performance. In current practice, network configuration depends heavily on field experience of engineers and often remains static for a long period of time. This practice is far from optimal. To address this limitation, online-learning-based approaches have great potentials to automate and optimize network configuration.  Learning-based approaches face the challenges of learning a highly complex function for each base station and balancing the fundamental exploration-exploitation tradeoff while minimizing the exploration cost. Fortunately, in cellular networks, base stations (BSs) often have similarities even though they are not identical. To leverage such similarities, we propose  kernel-based multi-BS contextual bandit algorithm based on multi-task learning. In the algorithm, we leverage the similarity among different BSs defined by conditional kernel embedding. 
We present theoretical analysis of the proposed algorithm in terms of regret and multi-task-learning efficiency. We evaluate the effectiveness of our algorithm based on a simulator built by real traces.

\end{abstract}



\maketitle

\section{Introduction}
\label{se:introduction}



\begin{figure}[thbp]
\includegraphics[width=0.9 \linewidth]{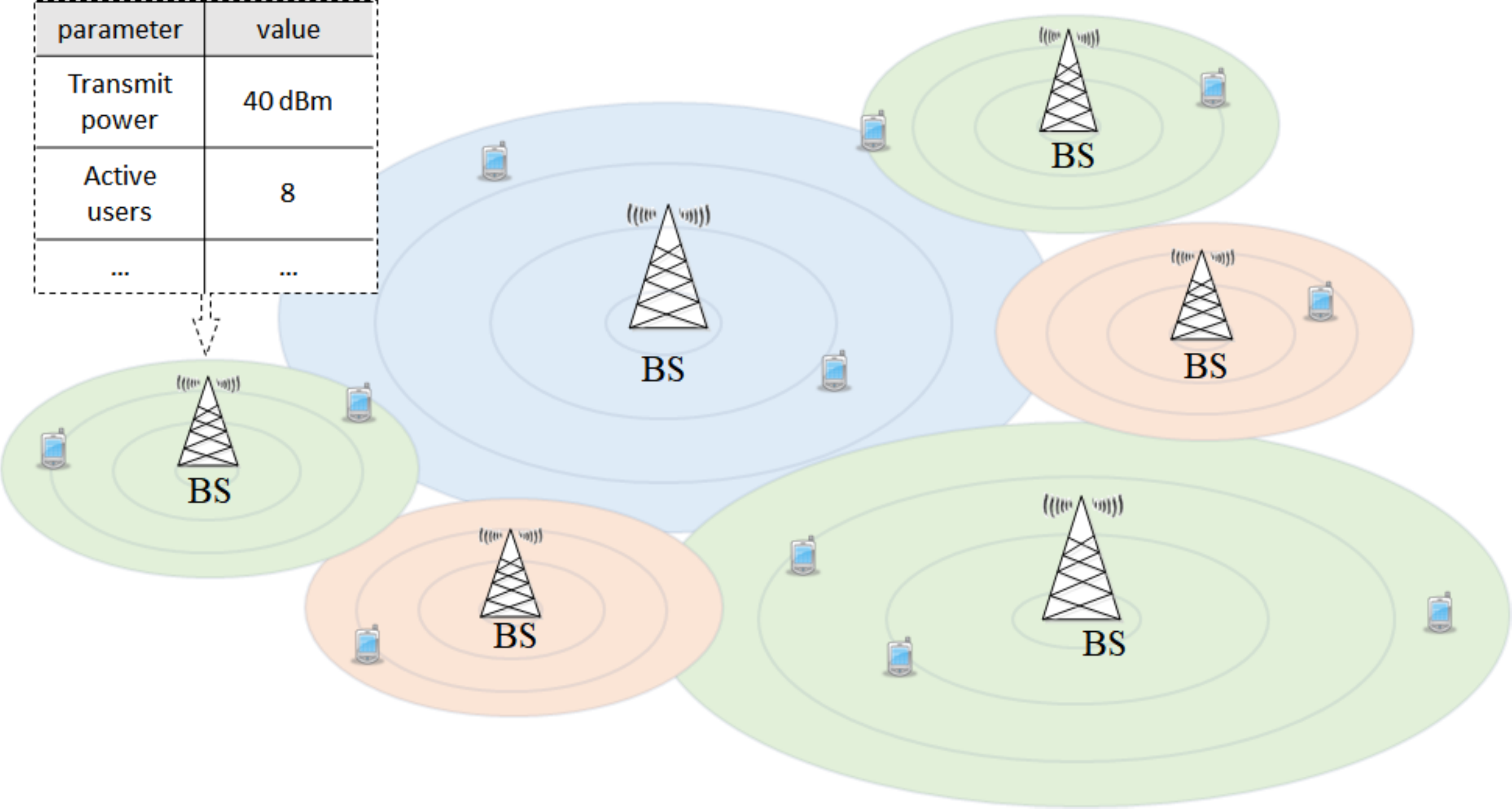}
\caption{Cellular network}
\label{fig:cellular network}
\end{figure}

With the development of mobile Internet and the rising number of smart phones, recent years have witnessed a significant growth in  mobile data traffic ~\cite{index2015cisco}. To satisfy the increasing traffic  demand, cellular providers are facing increasing pressure to further optimize their networks. Along this line, one critical aspect is  cellular  base station (BS) configuration. In cellular networks, a BS is a piece of network equipment that provides service to mobile users in its geographical coverage area (similar to a WiFi access point, but much more complex), as shown in Figure~\ref{fig:cellular network}. Each BS has a large number of parameters to configure, such as spectrum band, power configuration, antenna setting, and user hand-off threshold. These parameters have a significant impact on the overall cellular network performance, such as user throughput or delay. For instance, the transmit power of a BS  determines its  coverage and affects the throughput of all users it serves.

In  current practice,   cellular configuration needs manual adjustment and is mostly decided based on the field experience of engineers . Network configuration parameters typically remain static for a long period of time, even years, unless severe performance problems arise. 
This is clearly not optimal in terms of network performance: different base stations have different deployment environments (e.g., geographical areas), and the conditions of each BS (e.g., the number of users) also change over time. Therefore, as shown in Figure \ref{fig:cellular network}, setting appropriate parameters for each deployed BS based on its specific conditions  could significantly help the industry to optimize its networks. \ A natural way of achieving this goal is to apply online-learning-based algorithms in order to automate and optimize network configuration. 

Online-learning-based cellular BS configuration faces multiple challenges. First, the mapping between network configuration and performance is highly complex. Since different BSs have different deployment environments, they have different mappings between network configuration and performance, given a BS condition. Furthermore, for a given BS, its condition also changes over time  due to network dynamics, leading to different optimal configurations at different points in time. In addition, for a given BS and given condition, the impact of network configuration on performance is too complicated to model using   white-box analysis  due to the complexity and dynamics of  network environment, user diversity, traffic demand, mobility, etc.  
Second, to learn this mapping and to optimize the network performance over a period of time, operators face a fundamental exploitation-exploration tradeoff: in this case, exploitation means  to use the best known  configuration that benefits immediate performance but may overlook better configurations that are unknown; and exploration means to experiment with unknown or uncertain configurations which may have a better performance in the long run, at the risk of a potentially lower immediate performance. 
Furthermore, running experiments in cellular networks is disruptive  - users suffer poor performance under poor configurations. Thus, providers are often conservative when running experiments and would prefer  to reduce the number  of explorations needed in each BS.  Fortunately, in a cellular network, BSs usually have similarities, even though  they are not identical. Therefore, it would be desirable to effectively leverage data from different BSs by exploiting such  similarities.
 \begin{figure}[thbp]
\includegraphics[width=\linewidth]{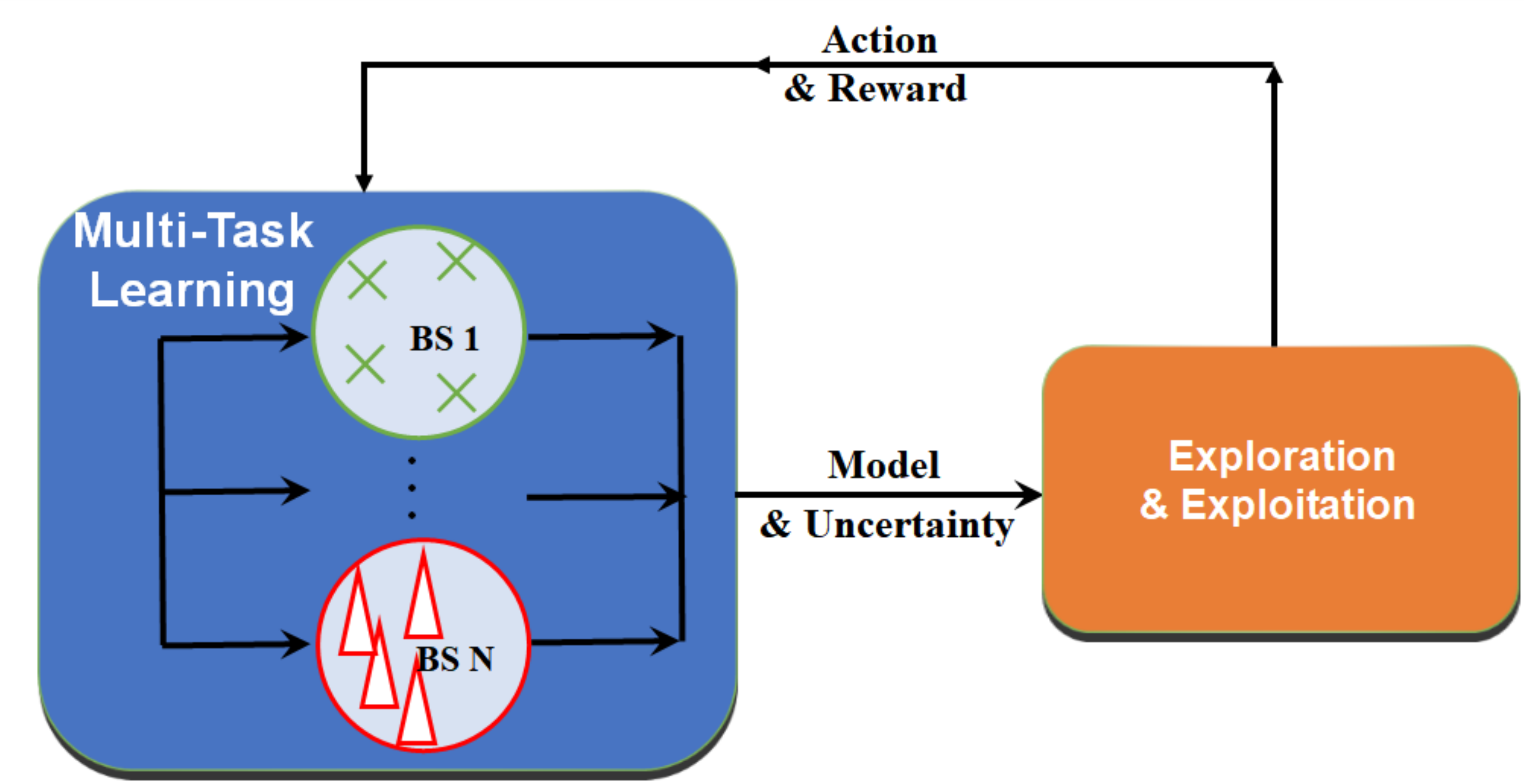}
\caption{Multi-task online learning }
\label{fig:framwork}
\end{figure}
 
To address these  challenges, we consider multiple BSs jointly and formulate the corresponding configuration problem as a multi-task on-line learning framework as shown in Figure \ref{fig:framwork}. The key idea is to leverage information from multiple BSs to jointly learn a model that maps the network state and its configuration to performance. The model is then customized to each BS based on its characteristics. Furthermore, the model also allows the BSs to  balance the tradeoff between the exploration and exploitation of the different configuration. Specifically, we propose a kernel-based multi-BS contextual bandits algorithm  that can leverage  similarity among BSs to automate and optimize cellular network configuration of multiple BSs simultaneously. Our contributions are multi-fold:

\begin{itemize}
	\item We develop a kernel-based multi-task contextual bandits algorithm to optimize cellular network configuration. The key idea is to explore similarities among BSs to make intelligent decisions about networks configurations in a sequential manner.
	\item We propose a method to estimate the similarity among the BSs based on the conditional kernel embedding.
    \item We present  theoretical guarantees  for the proposed algorithm in terms of regret and multi-task-learning efficiency.
	\item We evaluate our algorithm both in synthetic  data and real traces data and outperforms bandits algorithms not using multi-task learning by respectively up to $70.8\%$ and $64.8\%$ .

\end{itemize}
The rest of the paper is organized as follows.
The related work is in Sec.~\ref{se:related}.
We introduce the system model and problem formulation in Sec.~\ref{se:system}.
We present a kernel-based multi-BS contextual bandit algorithm  in Sec.~\ref{se:method}. The theoretical analysis of the algorithm is in Sec.\ref{se:analysis}.
We demonstrate the numerical results in Sec.~\ref{se:real}, 
and conclude in Sec.~\ref{se:conclusion}.
\section{Related Work}\label{se:related}
\textbf{Cellular Network Configuration} 
Various aspects of network parameter configuration have been studied in the literature, such as pilot power configuration, spectrum, handoff threshold, etc.
Traditional approaches derive analytical relationship between network configuration and its performance based on communication theory, such as \cite{Ding2012decomposition,Valkealahti2002wcdma,Ashraf2010distributed,Guo2016}. Such approaches are often prohibitively complex, involve various approximations, and require a significant amount of input information (such as the number of users, the location of each user, etc.). 

Recently, learning-based methods are proposed \cite{Razavi2011self,guo2017cellular,shen2018generalized,Chen2015}.
In \cite{Razavi2011self}, the authors propose a tailored form of reinforcement learning to adaptively select the optimal antenna configuration in a time-varying environment. 
In \cite{Chen2015}, the authors use Q-learning with compact state representation for traffic offloading. 
In \cite{shen2018generalized}, the authors design a generalized global bandit algorithm to control the  transmit power in the cellular coverage optimization problem.
In all these papers, BS similarities are not considered, and thus require more exploration. 
In \cite{guo2017cellular}, the authors study the pilot power configuration problem and design a Gibbs-sampling-based online learning algorithm so as to maximize the throughput of users. In comparison, they make the assumption that all BSs are equal while we allow different BSs to learn different mappings.

\textbf{Contextual Bandits} Contextual bandit \cite{langford2008epoch} is an extension of classic multi-armed bandit (MAB) problem \cite{auer2002finite}. One type of algorithm is the UCB-type such as Lin-UCB \cite{li2010contextual}, Kernel-UCB \cite{valko2013finite}, in which they assume the reward is a function of the context and trade off between the exploitation and exploration based on upper confident bound of the estimation \cite{auer2002using}. The contextual bandit is also widely used in many application areas, such as news article recommendation \cite{li2010contextual}, clinical trials \cite{villar2015multi}.

\textbf{Multi-task Learning} Multi-task learning has been extensively studied in many machine learning lectures\cite{evgeniou2004regularized}.
A common way is using a kernel function to define the similarity among tasks, e.g., in \cite{bonilla2008multi,bonilla2007kernel,deshmukh2017multi}. In \cite{deshmukh2017multi}, the authors design an algorithm that can \textbf{transfer information among arms} in the contextual bandit. Compared with \cite{deshmukh2017multi}, in our problem, we define an individual contextual bandit problem for each BS and consider the \textbf{multi-task learning among different contextual bandit problems}.
\section{System Model and Problem Formulation}
\label{se:system}
In this section, firstly, we describe the detail of the multi-BS configuration problem. Then we formulate the problem as a multi-task contextual bandits model. 
\subsection{Multi-BS Configuration}
We focus on the multi-BS network configuration problem. Specifically, we consider a set of BSs  $\mathcal{M}:=\{1,\cdots,M\}$  in a network. The time of the system is discretized, over a time horizon of $T$ slots. At time slot $t$, $\forall t \in \mathcal{T}:=\{1,\cdots,T\}$, for each BS $m \in \mathcal{M}$, its state is represented by a vector $s_t^{(m)}\in \mathbb{R}^d $. The state may include the number of users in a BS, user mobility, traffic demand, and neighboring BS configuration. 
At the beginning of each time slot $t$, the BS observes its state  $s_t^{(m)}$, and chooses its configuration $c_{t}^{(m)} \in \mathcal{C}_{action} \subset \mathbb{R}$ using a network configuration algorithm, where ${C}_{action}$ is a finite set.  $c_{t}^{(m)}$, ${C}_{action}$ are also respectively refer to as action and action space for consistency with reinforcement learning literature.
At the end of time slot $t$, the BS receives a resulting reward $r_{c_{t},t}^{(m)} \in \mathbb{R}$, which is a measure of network performance.
We note that $c_{t}^{(m)}$ can depend on all historical information of all BSs and the current state $s_t^{(m)}$.

In practice, the configuration  parameters can include pilot power, antenna direction, handoff threshold, etc.  The reward can be metrics of network performance, such as uplink throughput, downlink throughput, and quality-of-service scores.
Time granularity of the system is decided by network operators.  In the current practice, configurations can be updated daily  during midnight maintenance hours. To further improve network performance, network operators are moving towards more frequent network configuration updates, e.g., on an hourly basis, based on network states.  

The goal of the problem is to find the configuration $c_{t}^{(m)}$, for all $t$ and $m$ that maximizes the total cumulative  reward over time, i.e.,
\begin{equation}
\label{equ: goal}
\max_{c_{t}^{(m)}  \in \mathcal{C}_{action},\forall t} \sum_{m=1}^M \sum_{t=1}^T r_{c_{t},t}^{(m)}
\end{equation}
In this problem, for a given BS and a given state, we do not have {\em{a prior}} knowledge of  the reward  of its action. We need to learn such a mapping during the time horizon. In other words, when choosing 
$c_t^{(m)}$, one should consider  the historical information of all BSs and current state, i.e., $s_\tau^{(m)},c_\tau^{(m)},r_{c_\tau,\tau}^{(m)},\forall \tau <t, \forall m $ and $s_t^{(m)}$. The choice of $c_t^{(m)}$ and the corresponding reward also affect  future actions.  Therefore, there exists a fundamental exploitation-exploration tradeoff: exploitation is to use the best learned configuration that benefits the immediate reward but may overlook better configurations that are unknown; and exploration is to experiment unknown or uncertain configurations which may have a better reward in the long run, at the risk of a potentially lower immediate reward. 

Furthermore, we note that the action of one BS can be affected by the information of other BSs. Therefore, the information from multiple BSs should be leveraged jointly to optimize the problem in \eqref{equ: goal}. Also, note that the BSs are similar but not identical. Therefore, the similarity of BSs need to be explored and leveraged to optimize  the network configuration.

 


In summary, the goal of multi-BS configuration problem is to choose appropriate  actions for all time slot and all BSs to maximize the the problem defined  in Eq.~\eqref{equ: goal}.

\subsection{Multi-Task Contextual Bandit}

Multi-armed bandit (MAB)~\cite{auer2002finite} is a powerful tool in a sequential decision making scenario where at each time step, a learning task pulls one of the arms and observes an instantaneous reward that is independently and identically (i.i.d.) drawn from a fixed but unknown distribution. The task's objective is to maximize its cumulative reward by balancing the exploitation of those arms that have yielded high rewards in the past and the exploration of new arms that have not been tried. The contextual bandit model ~\cite{langford2008epoch} is an extension of the MAB in which each arm is associated with side information, called the context. The distribution of rewards for each arm is related to the associated context. The task is to learn the arm selection strategy by leveraging the contexts to predict the expected reward of each arm. Specifically, in the contextual bandit, over a time horizon of $T$ slots, at each time $t$, environment reveals context $x_{a_t,t} \in \mathcal{X}$ from set $\mathcal{X}$ of contexts for each arm $a \in \mathcal{A}$ from arms set $\mathcal{A}:=\{1,2,\cdots, N\}$, the leaner required to select one arm $a_t$ and then receives a reward $r_{a_t,t}$ from environment. At the end of the time slot $t$, learner improves arm selection strategy based on new observation $\{x_{a_t,t},  r_{a_t,t}\}$. At time $t$, the best arm is defined as $a_t^{*} = \arg \max_{a\in \mathcal {A}} \mathbb{E}(r_{a_t,t}^{}|x_{a_t,t}^{})$ and the corresponding reward is $r_{a_t^{*},t}^{}$. The regret at time $T$ is defined as the sum of the gap between the real reward and the optimal reward through the $T$ time slots in Eq. (\ref{equ: regret}). The goal of maximization of the accumulative reward $\sum_{t=1}^{T}r_{a_t,t}$ is equivalent to minimizing the regret\footnote{This is pseudo regret \cite{bubeck2012regret}.}.
 \begin{equation}
 \label {equ: regret}{}
R(T) = \sum_{t=1}^T (r_{a_t^{*},t}^{}-r_{a_t,t}^{})  
 \end{equation}

Based on the classical contextual bandit problem, we propose a multi-task contextual bandit model. Consider a set of tasks $\mathcal{M}:=\{1,\cdots,M\}$ , each task  $m \in \mathcal{M}$ can be seen as a standard contextual bandit problem. 
More specifically, in task $m$, at each time $t$, for each arm $a \in \mathcal{A}$,  there is an associated context vector $x_{a,t}^{(m)} \in \mathbb{R}^p$. If the arm $a_t^{(m)}$ is pulled as time $t$ for task $m$, it receives a reward $r_{a_t,t}^{(m)}$. The detail is shown in Problem 1. 
 \begin{algorithm}[htb]
\floatname{algorithm}{Problem}
\caption{Multi-Task Contextual Bandit}  
\label{prob:Contextual}  
\begin{algorithmic}[1] 
\FOR {$t=1$ \text{ to } $T$ }
\STATE Environment reveals  context $x_{a,t}^{(m)} \in \mathcal{X}$ for each arm $a\in \mathcal{A}$ and each task $m \in \mathcal{M}$
\FOR {$ \forall m \in \mathcal{M}$ }
\STATE Selects and pulls an arm $a_t^{(m)} \in \mathcal{A}$ 
\STATE Environment reveals a reward $r_{a_t,t}^{(m)}\in [0,1]$
\ENDFOR
\STATE Improves arm selection  based on new observations $\{(x_{a_t,t}^{(m)},  r_{a_t,t}^{(m)})|m \in \mathcal{M}\}$
\ENDFOR
\end{algorithmic}  
\end{algorithm}
 
We also define the best arm as
 $a_t^{(m)*} = \arg \max_{a\in \mathcal {A}} \mathbb{E}(r_{a_t,t}^{(m)}|x_{a_t,t}^{(m)})$ and the corresponding reward is $r_{a_t^{*},t}^{(m)}$. The regret over time horizon $T$ is defined as the sum of the gap between the real reward and the optimal reward through the $T$ time slot among all $M$ tasks in Eq. (\ref{equ: mregret}). The goal of the problem is to minimize the regret.
\begin{align}
\label{equ: mregret}
R(T) = \sum_{m=1}^M \sum_{t=1}^T \left(r_{a_t^{*},t}^{(m)}-r_{a_t,t}^{(m)}\right)
\end{align}

We can formulate the multi-BSs configuration problem as multi-task contextual bandit. \textbf{We regard the configuration optimization problem for one BS as one task.} Specifically, for each BS $m$, at time $t$, the context space $\mathcal{X}$ can be represented by a product of state space $\mathcal{S}$ and action space $\mathcal{C}_{action}$. And we index the finite set $\mathcal{C}_{action}$ by arms set $\mathcal{A}$, i.e., use $c_{a,t}$ to represent each possible configuration in time $t$. Then we define \textbf{context associated with arm $a$ is the combination of the state and the configuration, i.e., $x_{a,t}^{(m)} =(s_t^{(m)},c_{a,t}^{(m)})$, where $s_t^{(m)} \in \mathcal{S}$ and $c_{a,t}^{(m)} \in \mathcal{C}_{action}$.} Then the goal of finding the best arms which can maximize the total accumulative reward in Eq. (\ref{equ: goal}) is equivalent to minimizing the regret defined in Eq. (\ref{equ: mregret}).
\section{Methodology}
\label{se:method}
Most existing work on the contextual bandit problems, such as LinUCB \cite{li2010contextual}, KernelUCB \cite{valko2013finite} assume the reward is a function of the context, i.e., $r_{a_t,t} = f(x_{a_t,t})$. At each time slot $t$, these algorithms use the estimated function $\hat{f}(\cdot)$ to predict the reward of each arm according to the context at time $t$, i.e.,$\{x_{a_t,t}\}_{a\in \mathcal{A}}$. Based on the value and uncertainty of the prediction, they calculate the upper confident bound (UCB) of each arm. Then they select the arm $a_t$ that has the maximum UCB value and then obtains a reward $r_{a_t,t}$. Last, they update the estimated function $\hat{f}(\cdot)$ by the new observation $(x_{a_t},r_{a_t,t}).$


In our multi-BS configuration problem defined in Eq. (\ref{equ: goal}), if we model every BS as an independent classical contextual bandit problem and use the existing algorithm to make its own decision, it would lose  information across BSs and thus is not efficient. Specifically, in the training process, it would learn a group of function $\{f^{(m)}|m \in \mathcal{M}\}$ independently and ignore the similarity among them. In practice, the BSs that are configured simultaneously have lots of similar characteristics, such as geographical location, leading to similar reward functions. Furthermore, in the real case, since the configuration parameters have a large impact on the network performance, the cost of experience is expensive. We need an approach to use the data effectively. So, motivated by this observation, we design the kernel-based multi-BS contextual bandits that can leverage the similarity information  and share experiences among BSs, i.e., tasks.

In this section, we propose a framework to solve the problem in Eq. (\ref{equ: mregret}). We start with the regression model. Then we describe how to incorporate it with multi-task learning. Next, we propose kernel-based multi-BS contextual bandits algorithm in Sec.\ref{sec:Multi-BS}. In the last, we discuss the details of task similarity for real data in Sec. \ref{sect:sim}.
\subsection{Kernel Ridge Regression}

\label{sect: regression}

For the network configure problem, we need to learn a model from historical data that can predict the reward $r_{a_t,t}$ from the context $x_{a_t,t}$.
There are two challenges. First, the learned model should capture the non-linear relation between the configuration parameters, state (context) and the network utility (reward) in complex scenarios. Second, since the learned model is used in the contextual bandit model, it needs to not only offer the mean estimate value of the prediction but also a confidence interval of the estimation that can describe the uncertainty of the prediction. This important feature is used later to trade off exploration and exploration in the bandit model. 

To address these two challenges, we use kernel ridge regression to learn the prediction model that can capture non-linear relation and provide an explicit form of the uncertainty of the prediction. Furthermore, intuitively, the kernel function can be regarded as a measure of similarity among data points. which makes it suitable for the multi-task learning into it in Sec. \ref{se:Multi-Task Learning}. Let us briefly describe the kernel regression model.

Kernel ridge regression is a powerful tool in supervised learning to characterize the non-linear relation between the target and feature. For a training data set $\{(x_i,y_i)\}_{i=1}^n$, kernel method assumes that there exists a feature mapping  $\phi(x): \mathcal{X} \rightarrow \mathcal{H}$ which can map data  into a feature space in which a linear relationship $y=\phi(x)^T\theta$ between $\phi(x)$ and $y$ can be observed, where $\theta$ is the parameter need to be trained.  The kernel function is defined as the inner product of two data vectors in the feature space. 
$k(x,x')=\phi(x)^T\phi(x'), \forall x,x' \in \mathcal{X}$.

The feature space $\mathcal{H}$ is a Hilbert space of functions $f: \mathcal{X} \rightarrow \mathbb{R}$ with inner product $k<\cdot,\cdot>$. It can be called as the associated reproducing kernel Hilbert space (RKHS) of $k$, notated by $\mathcal{H}_{k}$. The goal of kernel ridge regression is to find a function $f$ in the RKHS $\mathcal{H}$ that can minimize the mean squared error of all training data, as shown in Eq. (\ref{equ: redge}).
\begin{equation}
\label{equ: redge}
\hat{f} =\arg \limits \min _{f \in \mathcal{H}_{k}}\sum_{i=1}^n (f(x_i)-y_i)^2+ \lambda||f||^2_{\mathcal{H}_{k}}
\end{equation}
where $\lambda$ is the regularization parameter. Applying the representer theorem, the optimal $f$ can be represented as a linear combination of the data points in the feature space, $f(\cdot) = \sum_{i=1}^n \alpha_i k(x_i,\cdot)$. Then we can get the solution of Eq. (\ref{equ: redge})
\begin{equation}
\label{equ: redge_sol}
f(x) = \boldsymbol{k}_{{X}:x}^T(K+\lambda I)^{-1}\boldsymbol{y}
\end{equation}
where $\boldsymbol{y}=(y_1,\cdots,y_n)$, $K$ is the Gram matrix, i.e., $K_{ij}= k(x_i, x_j)$, $\boldsymbol{k}_{{X}:x}= (k(x_1,x),\cdots,k(x_n,x))$ is the vector of the kernel value between all historical data $X$ and the new data, $x$. 

This provides basis for our bandit algorithms. The uncertainty of  prediction of the kernel ridge regression is discussed in Sec.\ref{sec:Multi-BS}.
\subsection{Multi-Task Learning}
\label{se:Multi-Task Learning}
We next introduce how to integrate kernel ridge regression into multi-task learning which allows us to use similarities information among BSs.

In multi-task learning, the main question is how to efficiently use data from one task  to another task. Borrowing the idea from ~\cite{evgeniou2004regularized,deshmukh2017multi}, we define the regression functions in the followings:
\begin{equation}
f: \tilde{X} \rightarrow \mathcal{Y}
\end{equation}
where $\mathcal{\tilde{X}} = \mathcal{Z} \times\mathcal{X}$, $\mathcal{X}$ is the original context space, $\mathcal{Z}$ is the task similarity space,  $\mathcal{Y}$ is the reward space. For each context $x_{a_t,t}^{(m)}$ of BS $m$, we can associate it with the task/BS descriptor $z_m \in \mathcal{Z}$, and define $\tilde{x}_{a_t,t}^{(m)} = (z_m,x_{a_t,t}^{(m)})$ to be the augmented context. We define the following kernel function $\tilde{k}$ in (\ref{ak}) to capture the relation among tasks.
\begin{equation}
\label{ak}
\tilde{k}((z, x), (z', x'))  = k_{\mathcal{Z}} (z, z')k_{\mathcal{X}} (x, x')
\end{equation}
where $k_{\mathcal{X}}$ is the kernel defined in original context, and $k_{\mathcal{Z}}$ is the kernel defined in tasks that measures the similarity among tasks/BSs. Then we define the task/BS similarity matrix $K_{Z}$ as $(K_{Z})_{ij}=k_{\mathcal{Z}}(z_i,z_j)$. We discuss the training of this similarity kernel and similarity matrix in Sec.\ref{sect:sim}.

In the multi-tasks contextual bandit model, at time $t$, we need to train an arm selection strategy based on the history data we experienced, i.e.,$\{(x_\tau^{(m)},r_{a_\tau,\tau}^{(m)})|m \in \mathcal{M},\tau < t\}$. We formulate a regression problem in Eq. (\ref {eqa: regression_all})
\begin{equation}
\label{eqa: regression_all}
 \hat{f_t} =\arg \min _{f\in {\mathcal {H}_{\tilde{k}}}}\sum_{m=1}^M\sum _{\tau=1}^{t-1}(f(\tilde{x}_{{a}_{\tau},\tau}^{(m)})-r^{(m)}_{a,\tau})^2+\lambda ||f||_{\mathcal {H}_{\tilde{k}}}^{2}
\end{equation}
where $\tilde{x}_{a_\tau,\tau}^{(m)}$ is the augmented context of the arm $a_\tau$ for task $m$, which is defined as the combination of the task descriptor $z_m$ and origianl context $x_{a_t,t}^{(m)}$, i.e., $\tilde{x}_{a_t,t}^{(m)} = (z_m, x_{a_t,t}^{(m)})$.

Then we can get a similar result as Eq. (\ref {equ: redge_sol}) in Eq. (\ref {solution}) . The only difference is that we use the augmented context $\tilde{x}$ and new kernel $\tilde{k}$ instead of the $x$ and $k$.
\begin{equation}
\label{solution}
\hat{f_t}(\tilde{x}) = \tilde{\boldsymbol{k}}^T_{t-1}(\tilde{x})(\tilde{K}_{t-1}+\lambda I)^{-1}\boldsymbol{y}_{t-1}
\end{equation}
where $\tilde{K}_{t-1}$ is Gram matrix of $[\tilde{x}_{a_{\tau},\tau}^{(m)}]_{\tau<t,m \in \mathcal{M}}$, \\ $\tilde{\boldsymbol{k}}_{t-1}(\tilde{x}) =[\tilde{k}(\tilde{x},\tilde{x}_{a_{\tau},\tau}^{(m)})]_{\tau<t,m \in \mathcal{M}}$, and $\boldsymbol{y}_{t-1}= [r_{a_{\tau},\tau}^{(m)}]_{\tau<t,m \in \mathcal{M}}$.
For the hyper parameter of kernel $k_{\mathcal{X}}$ and  the regularization parameter $\lambda$, we can use maximum likelihood method to train them. Then we can use Eq.(\ref{solution}) to predict the network utility (reward) based on the  configured parameter and network state (augmented context). 
\subsection{Kernel-based Multi-BS Contextual Bandits}
\label{sec:Multi-BS}
Next, we introduce how to measure the uncertainty of the prediction in  Eq. (\ref{solution}). At time $T$, for a specific task, i.e., BS, $m \in \mathcal{M}$, for a given augmented context $\tilde{x}_{a_T,T}^{(m)}$ of an arm, in order to estimate the uncertainty of the prediction $\hat{f}_T(\tilde{x}_{a_T,T}^{(m)})$, we need to make an assumption that the reward at time $T$, i.e., $r_{a_T,T}^{(m)}$ and all historical reward data, i.e., $\{r_{a_\tau,\tau}^{(m)})|m \in \mathcal{M},\tau < T\}$ are all independent random variables. Then we can use McDiarmid's inequality to get an upper confident bound of the predicted value. Since the mathematical derivation of this step is the same as Lemma 1 in \cite{deshmukh2017multi}, we only make a minor modification to obtain Theorem \ref{upper}.
\begin{theorem}
\label{upper}
For task $\forall m \in \mathcal{M}$, suppose the rewards $r_{a_T,T}^{(m)}$ at time $T$ and the history reward $\{r_{a_\tau,\tau}^{(m)})|m \in \mathcal{M},\tau < T\}$  are independent random variables with means $E[r_{a_\tau,\tau}^{(m)}|\tilde{x}_{a_\tau,\tau}^{(m)}] = f^*(\tilde{x}_{a_\tau,\tau}^{(m)})$, where $f^* \in \mathcal{H}_{\tilde{k}}$ and $||f^*||_{\mathcal{H}_{\tilde{k}}}\leq c$. Let $\alpha = \sqrt{\frac{\log(2((T-1)MN+1)/\delta)}{2}}$ and $\delta > 0$. With probability at least $1- \frac{\delta}{T}$, we have that $ \forall a \in \mathcal{A}$
\begin{equation}
|\hat{f}_t(\tilde{x}_{a,t}^{(m)})-f^*(\tilde{x}_{a,t}^{(m)})|\leq (\alpha + c\sqrt{\lambda})\sigma_{a,t}^{(m)}
\end{equation}
where the width is 
\begin{align}
\label{eq:width}
\sigma_{a,t}^{(m)}\! =\! \sqrt{\tilde{k}(\tilde{x}_{a,t}^{(m)},\tilde{x}_{a,t}^{(m)})\! -\! \tilde{\boldsymbol{k}}_{t-1}^T(\tilde{x}_{a,t}^{(m)})(\tilde{K}_{t-1}+ \lambda I)^{-1}\tilde{\boldsymbol{k}}_{t-1}(\tilde{x}_{a,t}^{(m)})}
\end{align}
\end{theorem}
Based on Theorem \ref{upper}, we define the upper confident bound UCB for each arm for each task in Eq. (\ref{equ:ucb}),  where $\hat{f}_t$ is obtained from Eq. (\ref{solution}), and $\beta$ is a hyper parameter. 
\begin{align}
\label{equ:ucb}
\text{UCB}_{a,t}^{(m)} = \hat{f}_{t}(\tilde{x}_{a,t}^{(m)})+ \beta \sigma_{a,t}^{(m)} 
\end{align}
Then we propose Algorithm \ref{alg:group} to solve the multi-BS configuration problem. 
\setcounter{algorithm}{0}
\begin{algorithm}[htp]
\caption{Kernel-based multi-BS configuration}  
\label{alg:group}  
\begin{algorithmic}[1] 
\FOR {$t=1$ \text{ to } $T$ }
\STATE Update the Gram matrix $\tilde{K}_{t-1}$
\FOR {\text{all BS} $m\in \mathcal{M}$}
\STATE Observe system state at time $t$ for BS $m$: $s_t^{(m)}$ and determine the context feature $x_{a,t}^{(m)}$ for each $a \in \mathcal{A}$
\STATE Determine the task/BS descriptor $z_m$ and get the augmented context $\tilde{x}_{a,t}^{(m)}$
\FOR {\text{all arm } $a$ \text{ in } $\mathcal{A}$ \text{ at time } $t$}
\STATE $ucb_{a,t}^{(m)}=\hat{f}(x_{a,t}^{(m)})+ \beta \sigma_{a,t}^{(m)}$
\ENDFOR
\STATE For BS $m$, choose arm $a_t^{(m)}= \arg \max ucb_{a,t}^{(m)}$
\STATE Observe reward $r_{a_t,t}^{(m)}$
\ENDFOR
\STATE Update $y_t$ by $\{r_{a_t,t}^{(m)}|m \in \mathcal{M}\}$
\ENDFOR
\end{algorithmic}  
\end{algorithm}

In Algorithm \ref{alg:group}, at each time $t$, it updates the prediction model $\hat{f}_t$. Then for each task $m \in \mathcal{M}$, it uses the model to obtain the UCB of each arm $a \in \mathcal{A}$. Next it selects the arm that has the maximum UCB. Algorithm \ref{alg:group} can trade off between the exploitation and exploration in the multi-BS configuration problem. The intuition behind it is as following: if one configuration is only tried for few times or even yet tried, its corresponding arm's width defined in Eq.(\ref{eq:width}) is larger, which makes its UCB value larger, then this configuration will be tried in following time with high probability.

\textbf{Independent Assumption} 
Note that the independent assumption of Theorem \ref{upper} is not true in Algorithm \ref{alg:group}, because the previous rewards influence the arm selection strategy (prediction function), then influence the following reward. To address it, we select a subset of them to make this assumption hold true in Sec.~\ref{se:analysis}. 

\textbf{High Dimensionality}
In Algorithm \ref{alg:group}, it updates  $\tilde{K}_{t-1}$ in line 2 and recalculates $(\tilde{K}_{t-1}+ \lambda I)^{-1}$ in line 7 based on Eq. (\ref{solution}). Since at time $t$, the dimension of $\tilde{K}_{t-1}$ is $M(t-1)$ and the computation complexity of inverse it is $O(M^3t^3)$. It increases dramatically over time. To address this issue, we use the Schur complement \cite{zhang2006schur} as following to simplify it.
\begin{theorem}
For a  matrix $M = \left[
\begin{matrix}
 A&U\\
 V&C\\
\end{matrix}
\right] 
$, define Schur complement of block $C$ as $S := A-UC^{-1}V$. Then we can get
\begin{align}
M^{-1} &= \left[
\begin{matrix}
 A&U\\
 V&C\\
\end{matrix}
\right]^{-1} = \left[
\begin{matrix}
 S^{-1}&-S^{-1}UC^{-1}\\
 -C^{-1}VS^{-1}&C^{-1}VS^{-1}UC^{-1}+C^{-1}\\
\end{matrix}
\right]
\end{align}
\end{theorem}
Based on it, we can update $(\tilde{K}_{t}+ \lambda I)^{-1}$ by $(\tilde{K}_{t-1}+ \lambda I)^{-1}$. It decreases the computation complexity to $O(Mt^2).$ 

For the issue of dealing with large dimension of Gram matrix $K$  has been much studied in Chapter 8 of \cite{Rasmussen06gaussianprocesses}. Most of them are designed for the supervise learning cases. In our problem, based on thr feature of online learning, Schur complement method is more suitable and efficiency.


\subsection{Similarity}\label{sect:sim}
The kernel $k_{\mathcal{Z}} (z, z')$ that defines the similarities among the tasks/BSs plays a significant role in Algorithm \ref{alg:group}. When $k_{\mathcal{Z}} (z, z') = \mathbbm{1}{(m=m')}$, where $\mathbbm{1}$ is the characteristic function, Algorithm \ref{alg:group} is equivalent to running the contextual bandit independently for each BS. In this section, we discuss how to measure the similarity in real data if it is not provided. 

Suppose the ground truth function for task $i$ (i.e., BS $i$) is $y=f_i(x)$, we need to define the similarity among different BSs based on the ground truth functions $f_i(x)$. From a Bayesian view, $y=f_i(x)$ is equivalent to  the conditional distribution $P(Y_i|\mathcal{X}_i)$. Therefore, we can use the conditional kernel embedding to map the conditional distributions to operators in a high-dimensional space, and then define the similarity based on it. Let us start with the definition of kernel embedding and conditional kernel embedding.

\subsubsection{\textbf{Conditional kernel embedding}}
Kernel embedding is a method in which a probability is mapped to an element of a potentially infinite dimensional feature spaces, i.e., a reproducing kernel Hilbert space (RKHS) ~\cite{smola2007hilbert}. For a random variable in domain $\mathcal{X}$ with distribution $P(X)$ , suppose $k:\mathcal{X}\times \mathcal{X} \rightarrow \mathbb{R} $ is the positive definite kernels with corresponding RKHS $\mathcal{H}_{\mathcal{X}}$, 
the kernel embedding of a kernel $k$ for $X$ is defined as
\begin{equation}
\nu_x = \mathbb{E}_X[k(\cdot,x)]=\int k(\cdot,x)dP(x)
\end{equation}
It is an element in $\mathcal{H}_{\mathcal{X}}$.

For two random variable $X$ and $Y$, suppose $k:\mathcal{X}\times \mathcal{X} \rightarrow \mathbb{R} $ and $l:\mathcal{Y}\times \mathcal{Y} \rightarrow \mathbb{R} $ are  respectively the positive definite kernels with corresponding RKHS $\mathcal{H}_{\mathcal{X}}$ and $\mathcal{H}_{\mathcal{Y}}$. The kernel embedding for the marginal distribution $P(Y|X=x)$ is:
\begin{equation}
\nu_{Y|x} = \mathbb{E}_Y[l(\cdot,y)|x]=\int l(\cdot,y)dP(y|x)
\end{equation}
It is an element in $\mathcal{H}_{\mathcal{Y}}$. Then for the conditional probability $P(Y|X)$, the kernel embedding is defined as a conditional operator $O_{Y|X}:\mathcal{H}_{\mathcal{X}} \rightarrow \mathcal{H}_{\mathcal{Y}}$ that satisfies Eq. (\ref{equ:cdk}) 
\begin{equation}
\label{equ:cdk}
\nu_{Y|x} = O_{Y|X}k(x,\cdot)
\end{equation}
If we have a data set $\{x_i,y_i\}_{i=1}^n$, which are i.i.d drawn from $P(X,Y)$, the conditional kernel embedding operator can be estimated by:
\begin{equation}
\label{equ:cdke}
\hat{O}_{Y|X} = \Psi(K+\lambda I)^{-1}\Phi
\end{equation}
where $\Psi =(l(y_1,\cdot),\cdots,l(y_n,\cdot))$ and $\Phi =(k(x_1,\cdot),\cdots,k(x_n,\cdot))$ are implicitly formed feature matrix, $K$ is the Gram matrix of $x$, i.e., $(K)_{ij}=k(x_i,x_j) $

The definition of conditional kernel embedding provides a way to measure probability $P(Y|X)$ as an operator between the spaces $\mathcal{H}_{\mathcal{Y}}$ and $\mathcal{H}_{\mathcal{X}}$.

\subsubsection{\textbf{Similarity Calculation}}
In this section, we use the  conditional kernel embedding to define the similarity space $\mathcal{Z}$ and augmented context kernel $k_\mathcal{Z}$ in Eq. (\ref{ak}).

We define the task/BS similarity space as $\mathcal{Z} = P_{\mathcal{Y}|\mathcal{X}}$ , the set of all conditional probability distributions of $Y$ given $X$. Then for task/BS $m$, given a context $x_{a,t}^{(m)}$ for arm $a$ at $t$, we define the augmented context $\tilde{x}_{a,t}^{(m)}$ as $ (P_{\mathcal{Y}_m|\mathcal{X}_m},x_{a,t}^{(m)})$. 

Then we use the Gaussian-form kernel based on the conditional kernel embedding to define $k_\mathcal{Z}$:
\begin{equation}
\label{equ:sim}
k_{\mathcal{Z}}(P_{\mathcal{Y}_m|\mathcal{X}_m},P_{\mathcal{Y}_{m'}|\mathcal{X}_{m'}}) =\exp(-||O_{Y|X}^{(m)}-O_{Y|X}^{(m')})||^2/2\sigma_{Z}^2)
\end{equation}
where $||\cdot||$ is Frobenius norm, $O_{Y|X}$ is the conditional kernel embedding defined in Eq. (\ref{equ:cdk}) and can be estimated by Eq. (\ref{equ:cdke}). The hyper parameter $\sigma_{Z}$ can be heuristically estimated by the median of Frobenius norm of all dataset. In Eq. (\ref{equ:cdke}), it can only be used in explicit kernels. Next, we use the kernel trick to derive a form that does not include explicit features.

Given a group of data sets $D_m=\{(x_i,y_i)\}_{i=1}^{n_m}$, and  $k$ and $l$ are respectively two positive definite kernels with RKHS $\mathcal{H}_\mathcal{X}$ and $\mathcal{H}_\mathcal{Y}$, for data set $D_m$, we define $\Psi_m =(l(y_1,\cdot),\cdots,l(y_{n_m},\cdot))$ and $\Phi_m =(k(x_1,\cdot),\cdots,k(x_{n_m},\cdot))$ are implicitly formed feature matrix of $y$ and $x$. $K_m=\Phi_m^T\Phi_m$ and $L_m=\Psi_m^T\Psi_m$ are Gram matrix of all $x$ and $y$. and $\mathcal{O}_{Y|X}^{(m)}$ for  conditional kernel embedding.
According to Eq. (\ref{equ:cdke}), we have
\begin{align*}
\mathcal{O}_{Y|X}^{(m)} &= \Psi_m(K_{m}+\lambda I)^{-1}\Phi^T_m
\end{align*}
\begin{equation}
\label{equ:distance}
\begin{split}
||\mathcal{O}_{Y|X}^{(m)} - \mathcal{O}_{Y|X}^{(m')}||^2 =& tr(\mathcal{O}_{Y|X}^{(m)T}\mathcal{O}_{Y|X}^{(m)})-2tr(\mathcal{O}_{Y|X}^{(m)T}\mathcal{O}_{Y|X}^{(m')})\\
&+ tr(\mathcal{O}_{Y|X}^{(m')T}\mathcal{O}_{Y|X}^{(m')})
\end{split} 
\end{equation}
Define matrix $K_{mm'}$ and $L_{mm'}$ by $(K_{mm'})_{ij} = k(x_i,x_{j})$ and $(L_{12})_{ij} = l(y_i,y_j)$, where $(x_i,y_i)$ is the $i$-th data in $\mathcal{D}_m$ and $(x_j,y_j)$ is the $j$-th data in $\mathcal{D}_{m'}$, so as $K_{mm'}$ and $L_{m'm}$. Then for the second term in Eq. (\ref{equ:distance}),
\begin{align*}
tr(\mathcal{O}_{Y|X}^{(m)T}\mathcal{O}_{Y|X}^{(m')})
&= tr(\Psi_m(K_{m}+\lambda I)^{-1}\Phi^T_m\Phi_{m'}(K_{{m'}}+\lambda I)^{-1}\Psi_{m'}^T)\\
&= tr((K_{m}+\lambda I)^{-1}\Phi^T_m\Phi_m(K_{{m'}}+\lambda I)^{-1}\Psi_{m'}^T\Psi_m)\\
&= tr((K_{m}+\lambda I)^{-1}K_{m{m'}}(K_{{m'}}+\lambda I)^{-1}L_{{m'}m})
\end{align*}
After using the same trick for other terms, Eq. (\ref{equ:distance}) can be written as
\begin{equation}
\label{eq:closed_form}
\begin{aligned}
||\mathcal{O}_{Y|X}^{(m)} - \mathcal{O}_{Y|X}^{(m')}||^{2} 
&= tr((K_{m}+\lambda I)^{-1}K_{m}(K_{m}+\lambda I)^{-1}L_m)\\
&-{2}* tr((K_{m}+\lambda I)^{-1}K_{m{m'}}(K_{{m'}}+\lambda I)^{-1}L_{{m'}m}\\
&+tr((K_{{m'}}+\lambda I)^{-1}K_{{m'}}(K_{{m'}}+\lambda I)^{-1}L_{m'})
\end{aligned}
\end{equation}



Then we can use Eq. (\ref{eq:closed_form}) in Eq. (\ref{equ:sim}) to measure the similarity between tasks. We denote the conditional kernel embedding metric for measure similarity as CKE.

\subsubsection{\textbf{Other similarity metrics}}
\label{sec:sim}
In the above, the similarity is defined based on $P(Y|X)$ among tasks through conditional kernel embedding. In the practice, there are several ways to define similarity.

{\bf{The average $\mathbf{R^2}$ method}}:
For example, for data set  $\mathcal{D}_1$ and  $\mathcal{D}_2$, we can train a regression model on  $\mathcal{D}_1$ and test it on $\mathcal{D}_2$, then 
measure the similarity using the prediction accuracy. Specifically, in the test set, we can measure prediction through the coefficient of determination $R^2$ as,
\begin{align}\label{eq:coeff_accuracy}
R^2 = 1 - 
\frac{\eta_{\textrm{ss}}}{\eta_{\textrm{var}} M_{\textrm{sp}}},
\end{align}
where $\eta_{\textrm{ss}}$ is the sum of squared prediction errors, 
$\eta_{\textrm{var}}$ is the variance of the target, and $M_{\textrm{sp}}$ is the total number of samples.
The larger the value of $R^2$, the better the model can capture the observed outcomes.
Switch the training and testing data, we obtain another $R^2$. Then we can define the similarity base on the average of these two $R^2$.If the value is smaller than a negative threshold, we can define the similarity as 0.  

{\bf{The hyper parameter method}} In work  \cite{bonilla2008multi}, it regards the similarity as a covariance matrix among tasks, they train them with other hyper parameter in the model based on maximum likelihood metric. This method needs more computation resource.

In practice, using different similarity definitions may have different results. 
The selection of the method to define similarity is in general heuristic.
In this problem, we have conducted experiments using different similarity definitions in the evaluation section Sec.\ref{se:real}.
It turns out that conditional kernel embedding (CKE) has the best performance in this kernel-based multi-BS configuration problem, and thus described in detail in this section. 


\section{Theoretical Analysis}\label{se:analysis}
In this section, we provide theoretical analysis of Algorithm \ref{alg:group} based on the classical bandit analysis. The first part is about regret analysis and the second part is about the multi-task-learning efficiency.
\subsection{Regret Analysis}

In Algorithm  \ref{alg:group}, at each time slot $t$, it uses the trained model to make a decision for all BSs in parallel. This is not in the same form of classical bandit model. In order to  
simply the analysis, we make an sequential version in Algorithm \ref{alg:seq1}, in which at each time $t$, it receives the context (state and configuration) of one BS with its BS ID,  denoted by $V_t$, that is used to identify the BS index $m$. Then algorithm \ref{alg:seq1} obtains the augment context using $V_t$  and then makes a decision for the BS. In this manner, Algorithm \ref{alg:seq1} makes a decision for all BSs sequentially. The performance of parallel and sequential methods are similar when the number of BSs is moderate and all BSs come in order, as in our case, since
the difference of number of updates for the model in the parallel and sequential cases is small. It is also shown from the simulation that their performances are similar.

\begin{algorithm}[htp]
\caption{Sequential multi-BS configuration}  
\label{alg:seq1}  
\begin{algorithmic}[1] 
\FOR {$t=1$ \text{ to } $T$ }
\STATE Update the Gram matrix $\tilde{K}_{t-1}$
\STATE Observe the BS ID $V_t$ and the corresponding context features at time $t$: $x_{a,t}$ for each $a \in \mathcal{A}$
\STATE Determine the BS descriptor $z_m$ based on $V_t$ and get the augmented context $\tilde{x}_{a,t}$
\FOR {\text{all arm } $a$ \text{ in } $\mathcal{A}$ \text{ at time } $t$}
\STATE $ucb_{a,t}=\hat{f}(\tilde{x}_{a,t})+ \beta \sigma_{a,t}$
\ENDFOR
\STATE Choose arm $a_t= \arg \max ucb_{a,t}$ for BS $V_t$
\STATE Observe reward $r_{a_t,t}$
\STATE Update $y_t$ by $r_{a_t,t}$
\ENDFOR
\end{algorithmic}  
\end{algorithm}

 The regret of Algorithm \ref{alg:seq1} is defined by 
\begin{align}
R(T) =	\sum_{m=1}^M \sum_{t=1}^T (r_{a_t^*,t}^{(m)}-r_{a_t,t}^{(m)})\mathbbm{1}{(V_t=m)}
\end{align}

In Algorithm \ref{alg:seq1}, the estimated reward $\hat{r}_{a_t,t}$ at time $t$ can be regarded as the sum of  variables in history $[r_{a_\tau,\tau}]_{\tau<t}$ that are dependent random variables. It does not meet the assumption in Theorem \ref{upper}, thus we are unable to analysis the uncertainty of the prediction. 

To address this issue, as in \cite{auer2002using,auer2002finite}, we design the base version (Algorithm \ref{alg:baseseq1}) and super version (Algorithm \ref{alg:superseq1}) of Algorithm \ref{alg:seq1} in order to meet the requirement of Theorem \ref{upper}. These algorithms are only designed to help theoretical analysis. In Algorithm \ref{alg:superseq1}, it constructs special, mutually exclusive subsets $\{\Psi(s)\}_S$ of $t$s
the elapsed time to guarantee the event $\{t \in \Psi_{t+1}^{(s)}\}$ is independent of the rewards observed at times in $\Psi_t^{(s)}$. On each of these sets, it uses Algorithm \ref{alg:baseseq1}  as subroutine to obtain the estimated reward and width of the upper confident bound which is the same as Algorithm \ref{alg:seq1}.

The construction of  Algorithm \ref{alg:baseseq1} and  Algorithm \ref{alg:superseq1}  follow similar strategy of that in the proof of KernelUCB (see Theorem 1 in \cite{valko2013finite} or Theorem 1 in~\cite{deshmukh2017multi}). Then we can get the following theorem 3.
\begin{algorithm}[htp]
\caption{Base sequential multi-BS configuration}  
\label{alg:baseseq1}  
\begin{algorithmic}[1] 
\STATE \textbf{Input:} $\beta, \Psi \subset \{1,\cdots,t-1\}$ 
\STATE Calculate Gram matrix $\tilde{K}_{\Psi}$ and get $y_{\Psi}=[r_{a_\tau,\tau}]_{\tau \in {\Psi}}$
\STATE Observe the BS ID $V_t$ and corresponding context features at time $t$: $x_{a,t}$ for each $a \in \mathcal{A}$
\STATE Determine the BS descriptor $z_m$ and get the augmented context $\tilde{x}_{a,t}$
\FOR {\text{all arm } $a$ \text{ in } $\mathcal{A}$ \text{ at time } $t$}
\STATE $\sigma_{a,t} = \sqrt{\tilde{k}(\tilde{x}_{a,t},\tilde{x}_{a,t})-\tilde{k}_{a,\Psi}^T(\tilde{K}_{\Psi}+ \lambda I)\tilde{k}_{a,\Psi}}$
\STATE $ucb_{a,t}=\hat{f}(x_{a,t})+ \beta \sigma_{a,t}$
\ENDFOR
\end{algorithmic}  
\end{algorithm}
\begin{algorithm}[htp]
\caption{Super sequential multi-BS configuration}  
\label{alg:superseq1}  
\begin{algorithmic}[1] 
\STATE \textbf{Input:} $\beta, T \in \mathbb{N}$ 
\STATE Initialize $S  \leftarrow \log \lceil T \rceil$ and $\Psi_1^{(s)} \leftarrow \emptyset$ for all $s \in S$
\FOR {$t=1$ \text{ to } $T$}
\STATE $s \leftarrow 1$ and $\hat{\mathcal{A}}_1 \leftarrow \mathcal{A}$
\REPEAT
\STATE $\sigma_{a,t}, ucb_{a,t}$ for all $a \in \hat{\mathcal{A}}_{(s)}$ $\leftarrow$ BaseAlg($\Psi_t^{(s)},\beta$) 
\STATE $\omega_{a,t} = \beta \sigma_{a,t}$
\IF {$\omega_{a,t} \leq \frac{1}{\sqrt{T}}$ for all $a \in \hat{\mathcal{A}}_{(s)}$}
\STATE Choose $a_t = \arg \max_{a \in \hat{\mathcal{A}}_{(s)}} ucb_{a,t}$
\STATE $\Phi_{t+1}^{(s)} \leftarrow \Phi_{t}^{(s)}$ for all $s \in S$

\ELSIF {$\omega_{a,t} \leq 2^{-s}$ for all $a \in \hat{\mathcal{A}}_{(s)}$}
\STATE $\hat{\mathcal{A}}_{s+1} \leftarrow \{ a\in \hat{\mathcal{A}}_s| ucb_{a,t} \geq \max_{a' \in \hat {\mathcal{A}}_s} ucb_{a',t} -2^{1-q} \}$
\STATE $s \leftarrow s+1$
\ELSE
\STATE Choose $a_t \in \hat{\mathcal{A}}_{s}$ s.t. $\omega_{a_t,t} > 2^{-q}$
\STATE $\Phi_{t+1}^{(s)} \leftarrow \Phi_{t}^{(s)} \cup \{ t \}$ and $\forall s' \neq s, \Phi_{t+1}^{(s)} \leftarrow \Phi_{t}^{(s)}$
\ENDIF
\UNTIL $a_t$ is found
\STATE Observe reward $r_{a_t,t}$
\ENDFOR
\end{algorithmic}  
\end{algorithm}

\begin{theorem}
Assume that $r_{a,t}\in[0,1],\forall a \in \mathcal{A} , T\geq 1,||f^*||_{\mathcal{H}_{\tilde{k}}}\leq c_{\tilde{k}},\forall \tilde{x} \in \tilde{X}$ and tasks similarity matrix $K_Z$ is known. With probability $1-\delta$, the regret of Algorithm \ref{alg:superseq1} satisfies,
\begin{align}
\label{equ:regret_big_o}
\begin{split}
R(T)&\leq 2\sqrt{T}+10(\sqrt{\frac{\log(2TN(\log(T)+1)/\delta))}{2}}+c\sqrt{\lambda})\\
&\sqrt{2d\log(g([T])}\sqrt{T\lceil\log(T)\rceil}\\
&=O(\sqrt{T \log(g([T]))})
\end{split}
\end{align}
where $g([T])=\frac{det({\tilde{K}_{T+1}+\lambda I})}{\lambda^{T+1}}$ and $d=max(1,\frac{c_{\tilde{k}}}{\lambda})$
\end{theorem}
\subsection{Multi-task-learning Efficiency}
In this section, we discuss the benefits of multi-task learning from the theoretical view point.

In the sequential setting, i.e., Algorithm \ref{alg:seq1} and Algorithm \ref{alg:superseq1}, because all BSs/tasks come in order,  at time $t$, each task happens $n =\frac{t}{M}$ times. Let  $K_{X_t}$ be Gram matrix of $[x_{a_{\tau},\tau}^{(m)}]_{\tau \leq t,m \in \mathcal{M}}$ i.e., original context, $K_Z$ be the similarity matrix. 
Then, following Theorem 2 in \cite{deshmukh2017multi}, the following results hold,
\begin{theorem}
Define the rank of matrix $K_{X_{T+1}}$ as $r_x$ and the rank of matrix $K_Z$ as $r_z$. Then
\begin{equation*}
\log(g([T])) \leq r_z r_x \log\Bigg(\frac{(T+1)c_{\tilde{k}}+\lambda}{\lambda}\Bigg)
\end{equation*}
\end{theorem}

According to Eq. (\ref{equ:regret_big_o}), if the rank of similarity matrix is lower, which means all BSs/tasks have higher inter-task similarity, the regret bound is tighter.

We make the further assumption that all distinct tasks are similar to each other with task similarity equal to $\mu$. Define $g_{\mu}([T])$ as the corresponding value of $g([T])$ when all task similarity equal to $\mu$. 
According to Theorem 3 in \cite{deshmukh2017multi}, we have
\begin{theorem}
If $\mu_1 \leq \mu_2$,  then $g_{\mu_1}([T])\geq g_{\mu_2}([T])$
\end{theorem}
This shows that given the assumption that all tasks comes in order and number of tasks is fixed, when BSs/tasks are more similar, the regret bound of Algorithm \ref{alg:seq1} is tighter.
In our case, running all task independently is equivalent to setting the similarity as an identify matrix, i.e., $\mu =0$. So, based on the previous two theorems, we show the benefits of our algorithm using the multi-task learning.
\section{Evaluation}
\label{se:real}
In this section, we evaluate the performance of the proposed approach  Algorithm 1. and Algorithm 2. in both synthetic data and real network data.

\subsection{Synthetic data evaluation}
We use synthetic data to demonstrate the impact of similarity in multi-task regression. Thereafter, we test our algorithm performance based on synthetic data.
\subsubsection{\textbf{Similarity in regression}} We generate the reward function of tasks with pre-defined ground truth similarity based on Gaussian process. Then we train the regression model using different similarity and measure the performance of regression. In detail, we generate 2-task data sets in the following manner: (1) Each data set has 100 data points, $D_1 =\{x_i^1,y_i^1\}_{i=1}^{100}$ and $D_2 =\{x_i^2,y_i^2\}_{i=1}^{100}$, and each $x_i$ is randomly sampled from $ [0,1]\times[0,1]\subset \mathbb{R}^2$ and $y\in \mathbb{R}$. (2) The ground truth similarity between two tasks is  $sim_{g}=0.8$.
i.e., the similarity matrix $K_{Z}$ is a symmetric $2\times2$ matrix with 1s in the main diagonal and 0.8s in the anti-diagonal.
(3) The kernel of $x$ is the Gaussian kernel with lengthscale 0.5. (4) ${\mathbf y}=[y_1^1,y_2^1,\cdots,y_{100}^1,y_1^2,y_2^2,\cdots,y_{100}^2]^T$ 
is sampled from a multivariate normal distribution with zero mean and whose covariance is the Kronecker product of similarity matrix $K_{Z}$ and the Gram matrix of $x$, ${K_X}$ added  white noise, i.e., $\mathbf{y} \sim \mathcal{N}(\mathbf{0},K_{Z}\otimes K_{X}+\sigma_{noise}^2\mathbf{I})$ with $\sigma_{noise}^2=0.05$.  (5) We sampled $Y$ for 100 times, and test the regression for each sampled $Y$.  (6) For each task, the size of train set is 5, other 95 data points are test data.

\begin{figure}[thbp]
\includegraphics[width=0.8 \linewidth]{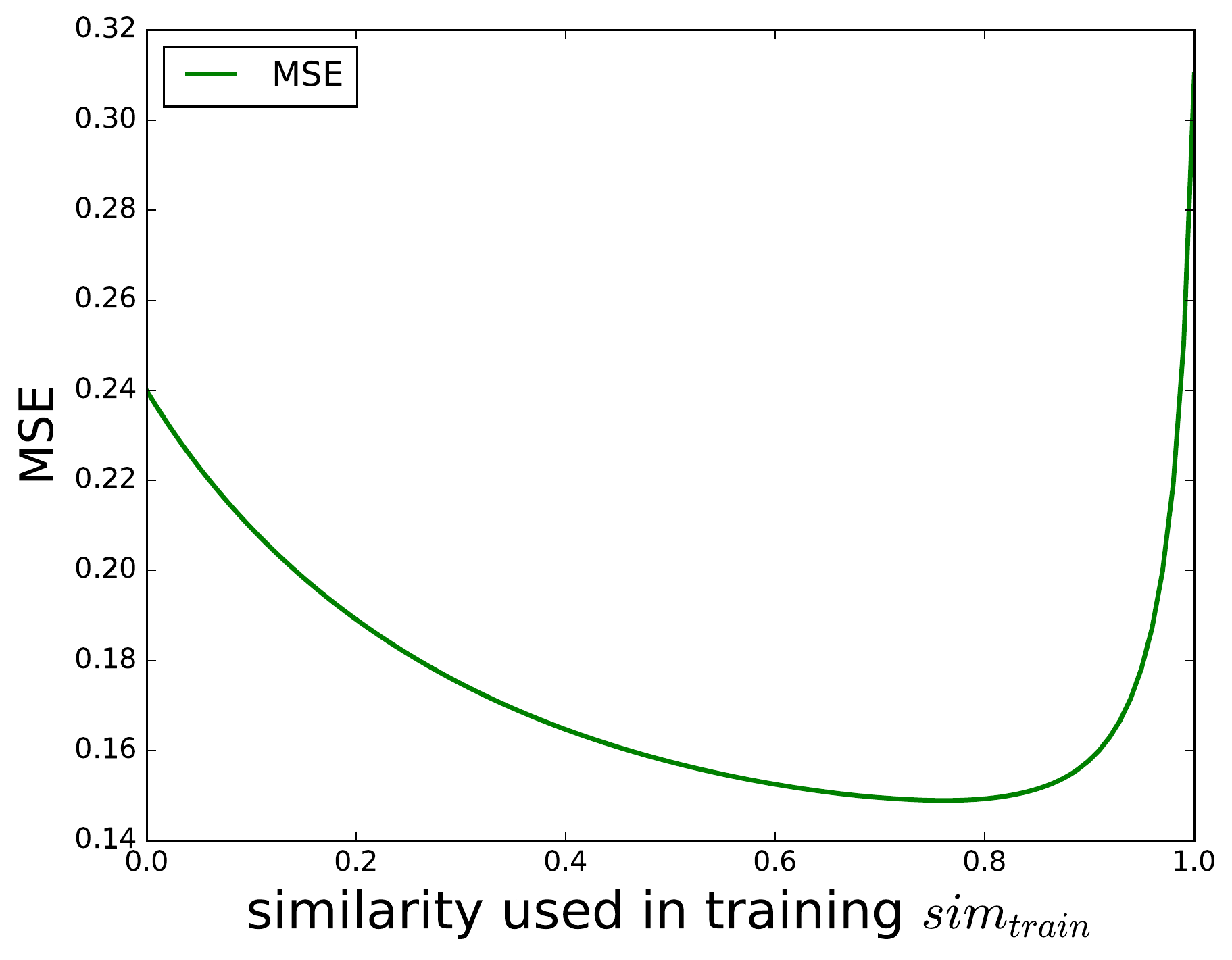}
\caption{Similarity v.s MSE in 2-task regression}
\label{fig:compare_sim}
\end{figure}

In the training process, the hyper parameter of the kernel are the same as the ones in the data generating process. 
For any similarity value  $sim_{train} \in [0,1]$ with granularity $0.01$ between two tasks, we use Eq. ({\ref{solution}}) to train the regression function. The performance is measured by mean square error (MSE) for all test data. 
The results is shown in Fig. \ref{fig:compare_sim}. 
The MSE is the the average of 100 samples  $\mathbf{y}$.  
It shows that the relation between MSE and similarity $sim_{train}$ is a convex form function. 
The case $sim_{train} =0$ is to train two tasks independently, that is, no information is shared between tasks; The case $sim_{train} =1$ is to train two tasks with the combination of the two data sets, that is, the difference between tasks is neglected.   
The best performance (minimum MSE) is achieved, when $sim_{train}=sum_g=0.8$, that is, similarity used in training is equal to the ground truth similarity. 
This is in accordance with our motivation to take the similarity measurement into the multi-task learning.    

\subsubsection{\textbf{Multi-task contextual bandit in synthetic data}} 

\begin{figure}[htp]
\includegraphics[width=0.8 \linewidth]{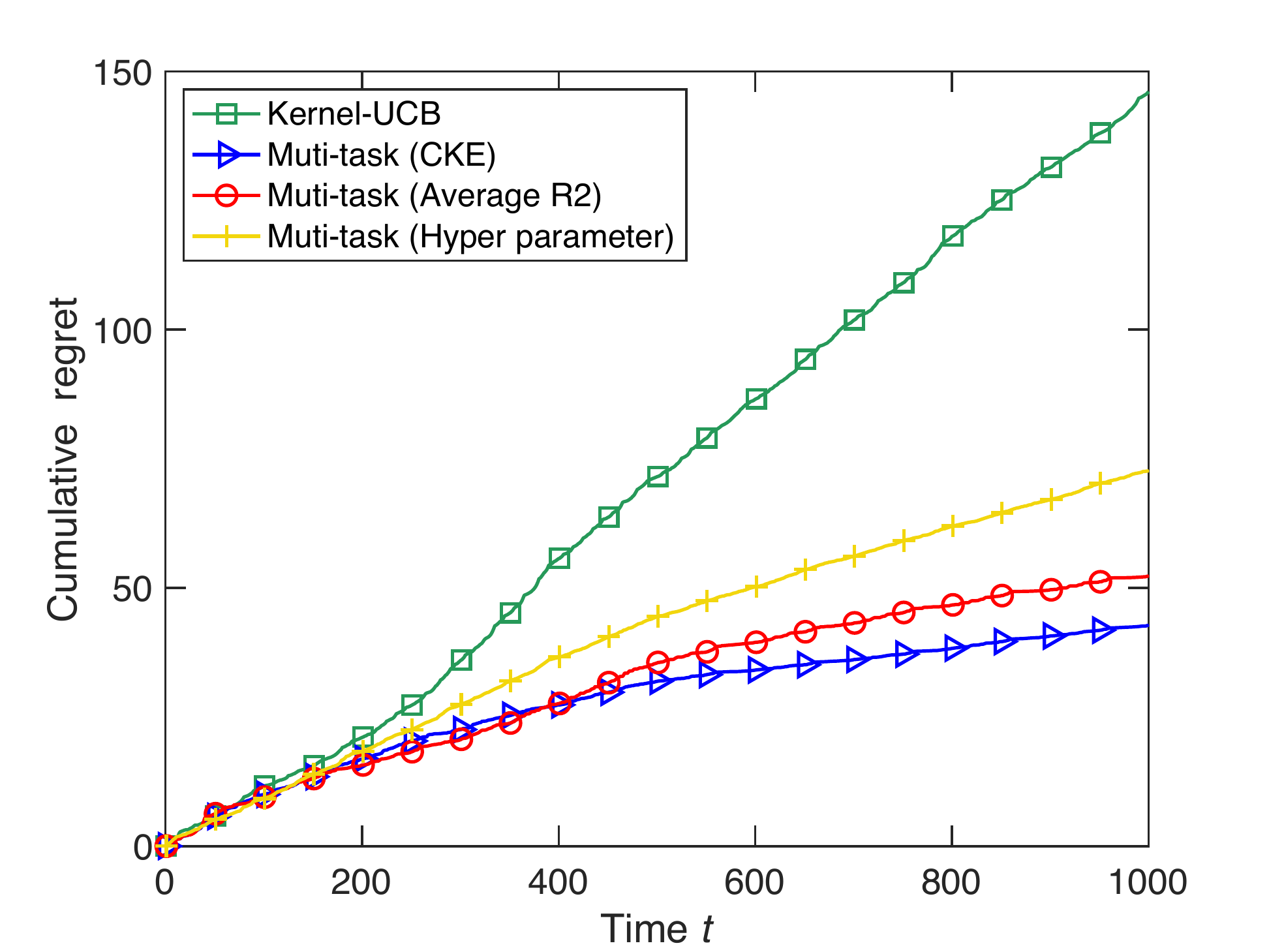}
\caption{Multi-task learning in synthetic data.}
\label{fig:syn}
\end{figure}

We use synthetic data to test the performance of  Algorithm \ref{alg:group} based on different similarity metrics in Sec. \ref{sect:sim}, the CKE, the average $R^2$ method and the hyper parameter method. Suppose that we have 5 tasks and 5 arms for each task, and define the context for each arm as $x_{a_t,t}^{(m)} \in \mathbb{R}^2$. To create the similar reward function for each task, we assume that there exits a hidden parameter $u_{t}$, which is randomly sampled from $[0,1]\times [0,1]\subset \mathbb{R}^2$, and the context for each arm $x_{a_t,t}^{(m)}$ is a projection of $u_{t}$, and the projection angle depends on the arm and task. Specifically, we use $u_{t}[0]$ and $u_{t}[1]$ to denote vector $u_{t}$'s first and second dimension. For task $m$, arm $a_t \in \mathcal{A} =\{1,2,3,4,5\}$, the corresponding  $x_{a_t,t}^{(m)} = [u_{t}[0] cos(\frac{\pi}{2}(\frac{a_t}{5}+\frac{m}{10})),u_{t}[1]sin(\frac{\pi}{2}\frac{a_t}{5})]$  and the reward is $r_{a,t}^{(m)} = 1-( u_t[0]-\frac{a_t}{5}+0.3-\frac{m}{10})^2$. We conduct the experiment in multi-task learning in parallel manner (same as Problem 1). The simulation result is  shown in Fig.\ref{fig:syn}.  
We compare the cumulative regret of  Algorithm \ref{alg:group}  with the performance of conducting Kernel-UCB  \cite{valko2013finite}  on each task independently. 
\begin{table*}
  \caption{Sample Data}
  \label{tab:data}
  \centering
  \begin{tabular}{cccp{0.21\columnwidth}p{0.24\columnwidth}p{0.2\columnwidth}p{0.2\columnwidth}p{0.38\columnwidth}}
    \toprule
    BS ID&\# Active users&\% CQI &\%Small packet SDUs&\%Small packet volume&\# Users&Threshold handover &\%Users throughput$\geq$5Mbps\\
    \midrule
    3714&	0.083643988	&0.342990&	61.37669801	&47.70435832	&5.20244	&-93&	90.78014184\\
    3714&	0.163259998	&0.606118&	35.45774141	&29.14181596&	7.89750&	-94	&82.55813953\\
    1217	&1.471931100	&0.242817&	30.86999337	&31.98075091&	85.12305&	-98	&84.06884082\\
    1217&	1.479040265	&0.437417&29.61262810	&21.28883741&	100.42472&	-101&	62.58613608\\
  \bottomrule
\end{tabular}
\end{table*}

Here, the cumulative regrets shown in Fig.\ref{fig:syn} are the sum of the cumulative regrets of the five tasks. Further, each data point is the average result of 10 individual simulations. It shows that the regret of multi-learning grows slower than the one of the Kernel-UCB.  After 1000 time slots, the multi-task learning (Algorithm \ref{alg:group}) using similarity base on CKE, the average $R^2$ and the hyper parameter method respectively decrease $70.8\%$, $64.2\%$ and $36.7\%$ of the regret compared to Kernel-UCB. 
We also test the sequential case (Algorithm \ref{alg:seq1}) in this setting, the performances are similar.

\subsection{\bf{Real data evaluation}}
 We start with the data collection and simulator construction procedure, and then discuss about the numerical results.
\subsubsection{\bf{Data Collection and Simulator Construction}} We build a network simulator based on data collected in real networks to provide interactive environment for bandit algorithms.

The data is collected in the real base station configuration experiments conducted  by a service provider in a metropolitan city. 
We employ 105 BSs within the region to collect
56580 data samples, each for the statistics of a BS observed from 2pm to 10pm in 5 days.  
An example is illustrated in Table  \ref{tab:data}. 
These statistics include network measurements,  and configured parameter.
The network measurements include user number, CQI, average packet size, etc, as illustrated from Column 2 to 6, used as {\bf states} in our experiment. 
The configured parameter is handover threshold, as shown Column 7, employed as {\bf actions/configurations}. 
To be specific, handover is a procedure for a BS to guarantee the user experience in cellular network. If one BS observes the signal strength of a user it serves is lower than the threshold, it will handover the user to another BS that has a better communication quality.  The range of the configured parameter values is from -112 dBm to -84 dBm, with 1 dBm resolution. 
Each base station change its configured parameter randomly several times per day. 
The {\bf reward} is 
the ratio of users with throughput no less than 5 Mbps, as shown in Column 8.

With the data, we build our simulator. 
The input is the state and configured parameter $(s,c_{a})$, and the output is the corresponding reward $r$. 
In detail, when the simulator receives the input $(s,c_{a})$, it returns the average of the rewards of the top $k$ nearest neighbors of $(s,c_{a})$ in the data set, by Euclidean distance. 

\subsubsection{\bf{Evaluation Setup and Results}}

\begin{figure}[hbp]
\begin{center}
\includegraphics[width=1.05\linewidth]{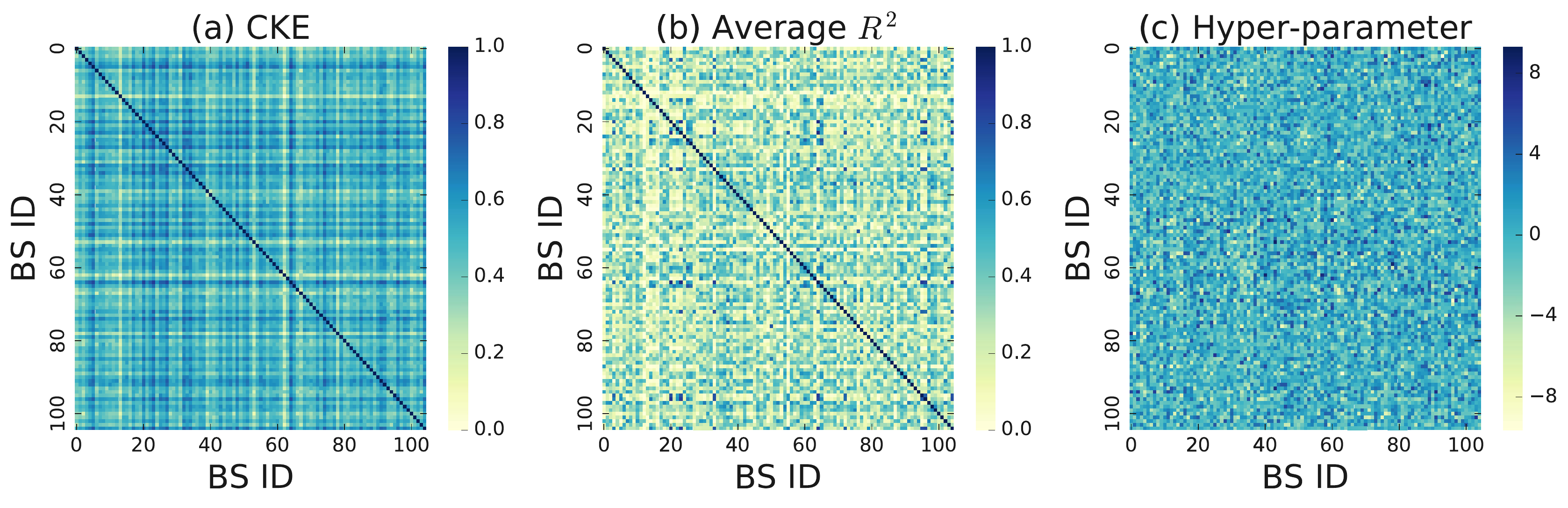}
\caption{Similarity matrix among 105 BSs}
\label{fig:sim}
\end{center}
\end{figure}

\begin{figure}[thbp]
\addtolength{\subfigcapskip}{-4pt}
\begin{center}
\captionsetup{justification=centering,margin=0.0cm}
\subfigure[Sequential case]{\includegraphics[angle = 0,width = 0.8\linewidth]{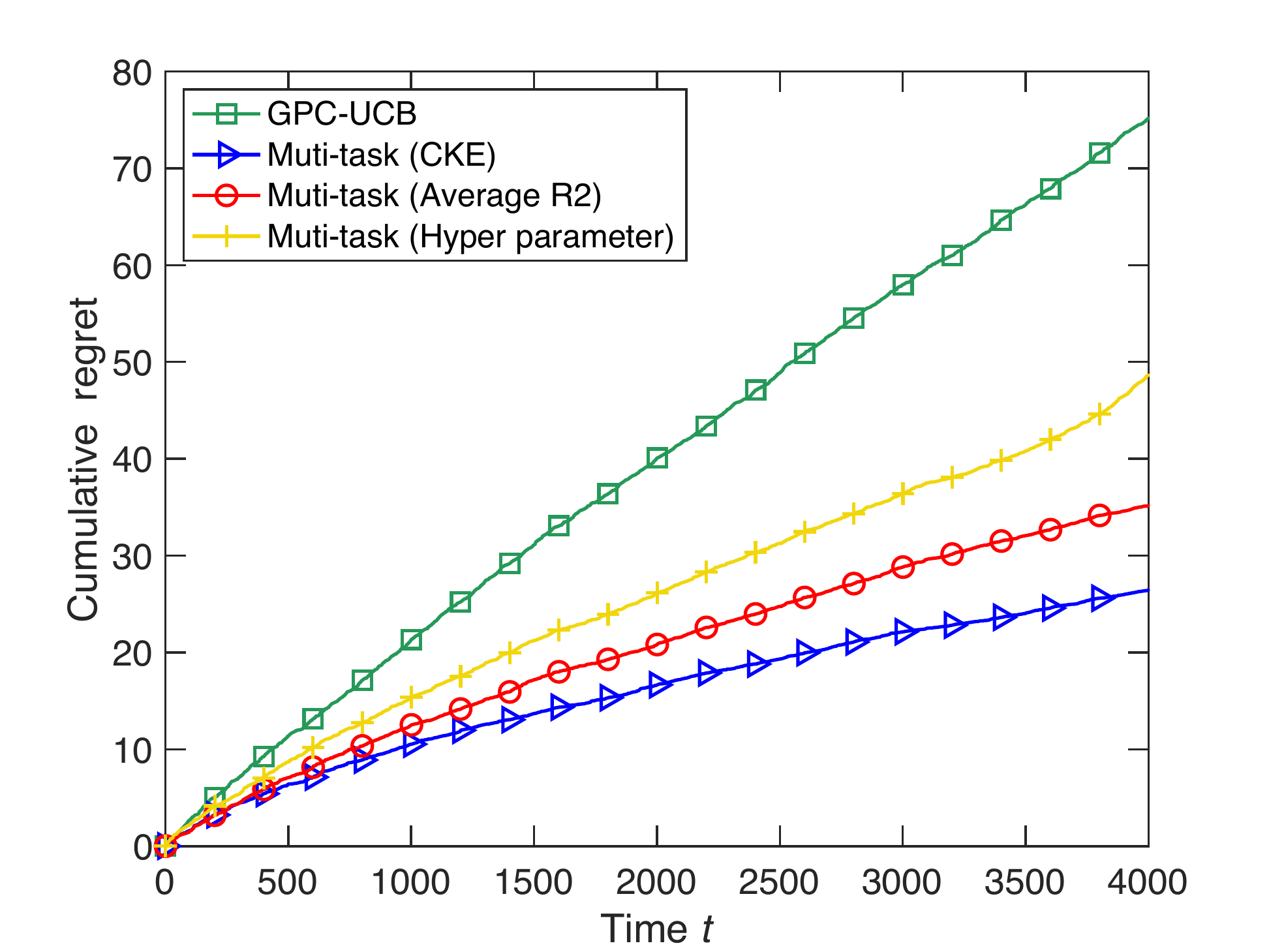}
\label{fig:sq}}
\subfigure[Parallel case]
{\includegraphics[angle = 0,width = 0.8\linewidth]{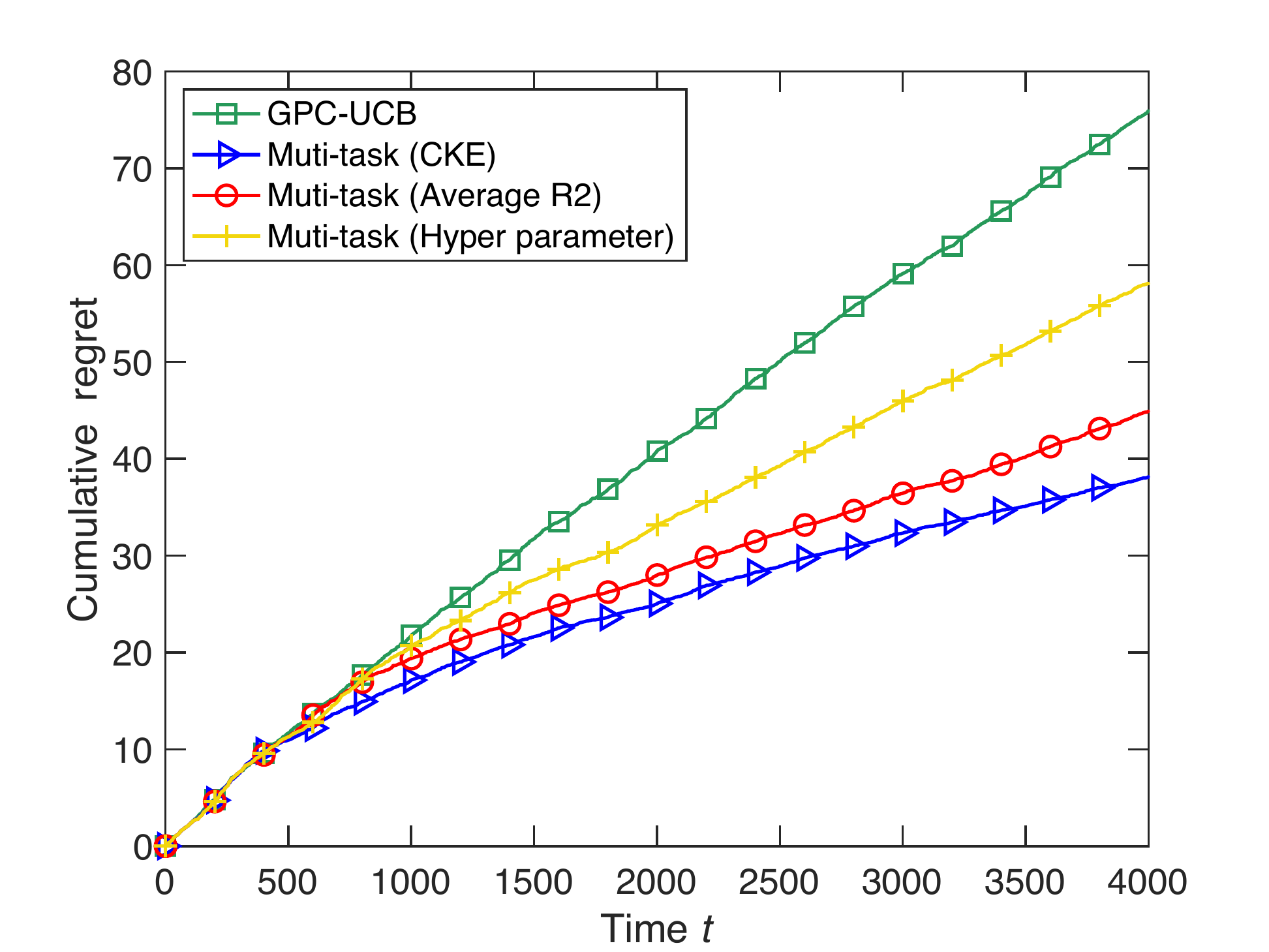}
\label{fig:pa}}
\caption{Multi-task learning v.s. Independent learning in real data}
\label{fig:real}
\end{center}
\end{figure}

In this experiment, the dimension of the state space is $5$. 
The action space $\mathcal{C}_{action}$is from -112 dBm to -84 dBm with 1 dBm resolution, that is, the number of arms in our model is $29$. 
The reward space is $[0,1]$. 
We test 3 methods to measure the similarity of 105 different BSs. 
In Fig. \ref{fig:sim}, each subplot corresponds to the similarity matrix $K_{Z}$ trained by methods in Sec. \ref{sect:sim}, the CKE, the average $R^2$ method and the hyper parameter method. 
The  value in Row $i$, Column $j$ corresponds to the similarity between BS $i$ and BS $j$. 
We test the multi-task learning case for all 105 BSs in the sequential and the parallel cases based on different similarity metrics. In Fig. \ref{fig:real} (a), the result for Algorithm \ref{alg:seq1} using similarity  matrix $K_Z$ in Fig. \ref{fig:sim} is shown. 
We compare the cumulative regret of  Algorithm \ref{alg:seq1} with the performance of conducting GPC-UCB  \cite{krause2011contextual} on each BS independently. 
To the best of our knowledge, GPC-UCB is the best algorithm acting on clear definitions of states and actions, therefore, we choose it as our baseline. 
The cumulative regrets shown in Fig.~\ref{fig:real}(a) are the sum of the cumulative regrets of the all BSs. 
Each data point is the average result of 10 individual simulations. 
It can be seen that, when our algorithm is used, the regret increases much slower than the baseline.  
In sequential case, after 4000 time slots, Algorithm \ref{alg:seq1}  using similarity base on the CKE, the average $R^2$ and the hyper parameter method decreases $64,8\%$, $53.2\%$ and $35.3\%$ of the regret compared to the baseline. 
For the parallel case, in  Fig. \ref{fig:real} (b), the result for Algorithm \ref{alg:group} using same similarity  matrix $K_Z$ is shown. 
To make a fair comparison, we rescale the time slots of the parallel case such that the size of the training data is the same as the one in the sequential case.  
In the parallel case, after 4000 time slots, Algorithm \ref{alg:group} using similarity based on the CKE, the average $R^2$ and  the hyper parameter method decreases $49.8\%$, $40.9\%$ and $23.5\%$ of the regret compared to the baseline. 
These figures show that the algorithm in the sequential case has better performance than the one in the parallel case. This is 
because the learning algorithm in sequential case can improve the model with the immediate feedback reward from each BS, 
while in the parallel case the algorithm only improves the model when all the feedback rewards from all BSs are collected. 

\section{Conclusion}\label{se:conclusion}

In this work, in order to address the multi-BS network configuration problem, we propose a kernel-based multi-task contextual bandits algorithm that  leverages the similarity among BSs effectively. In the algorithm, we also provide an approach to measure the similarity among tasks based on conditional kernel embedding. Furthermore, we present theoretical bounds for the proposed algorithm in terms of regret and multi-task-learning efficiency.  It shows that the bound of regret is tighter if the learning tasks are more similar. We also evaluate the effectiveness of our algorithm on the synthetic data and the real problem based on a simulator built by real traces. Future work includes possible experimental evaluations in real field tests and further studies on the impact of different similarity metrics.



\bibliographystyle{ACM-Reference-Format}
\bibliography{sample-bibliography}

\end{document}